\documentclass[a4paper, 10pt, twocolumn]{article}   
\usepackage{preprint}
        
\usepackage{cite}
\usepackage{color}

\usepackage{bm}

\usepackage{graphicx}
\usepackage{amsmath}
\usepackage{amssymb}
\usepackage{subcaption}

\usepackage{hyperref} 

\usepackage{multirow}

\usepackage{tikz}
\usetikzlibrary{arrows} 
\usetikzlibrary{calc} 
\usepackage{pgfplots}

\usepackage{nicefrac} 

\usepackage[capitalize]{cleveref} 

\usepackage{comment}

\title{
A Geometric Approach to On-road Motion Planning\\for Long and Multi-Body Heavy-Duty Vehicles
}

\author{Rui Oliveira$^{1, 3}$, Oskar Ljungqvist$^{2}$, Pedro F. Lima$^{3}$, Jonas M{\r{a}}rtensson$^{1}$, and Bo Wahlberg$^{1}$}

\begin{document}

\twocolumn[ 
  \begin{@twocolumnfalse} 
  
\maketitle
\begin{abstract}
Driving heavy-duty vehicles, such as buses and tractor-trailer vehicles, is a difficult task in comparison to passenger cars.
Most research on motion planning for autonomous vehicles has focused on passenger vehicles, and many unique challenges associated with heavy-duty vehicles remain open.
However, recent works have started to tackle the particular difficulties related to on-road motion planning for buses and tractor-trailer vehicles using numerical optimization approaches.
In this work, we propose a framework to design an optimization objective to be used in motion planners.
Based on geometric derivations, the method finds the optimal trade-off between the conflicting objectives of centering different axles of the vehicle in the lane.
For the buses, we consider the front and rear axles trade-off, whereas for articulated vehicles, we consider the tractor and trailer rear axles trade-off.
Our results show that the proposed design strategy results in planned paths that considerably improve the behavior of heavy-duty vehicles by keeping the whole vehicle body in the center of the lane.

\end{abstract}
\vspace{0.35cm}

  \end{@twocolumnfalse} 
] 

\newcommand\blfootnote[1]{%
  \begingroup
  \renewcommand\thefootnote{}\footnote{#1}%
  \addtocounter{footnote}{-1}%
  \endgroup
}
\newcommand\extrafootertext[1]{%
    \bgroup
    \renewcommand\thefootnote{\fnsymbol{footnote}}%
    \renewcommand\thempfootnote{\fnsymbol{mpfootnote}}%
    \footnotetext[0]{#1}%
    \egroup
}

\extrafootertext{$^{1}$Division of Decision and Control Systems, School of Electrical Engineering and Computer Science, KTH Royal Institute of Technology, Stockholm, Sweden
        {\tt\small rfoli@kth.se}, {\tt\small jonas1@kth.se}, {\tt\small bo@kth.se}}%
\extrafootertext{$^{2}$Division of Automatic Control, Link\"oping University, Link\"oping, Sweden
    	{\tt\small oskar.ljungqvist@liu.se}}%
\extrafootertext{$^{3}$Scania CV AB, Autonomous Transport Solutions, S{\"o}dert{\"a}lje, Sweden
		{\tt\small pedro.lima@scania.com}}%
\extrafootertext{*This work was partially supported by the Wallenberg AI, Autonomous Systems and Software Program (WASP) funded by the Knut and Alice Wallenberg Foundation.}%

\def\DTra{L_2}
\def\DHit{M_1}

\def\BusSuperscript{bus}
\def\BusFrontAxleEy{e_y^{\BusSuperscript}}
\def\BusFrontAxleEyBar{{\bar e}_y^{\BusSuperscript}}
\def\BusFrontAxleEyHat{{\hat e}_y^{\BusSuperscript}}

\mathchardef\mhyphen="2D
\def\TrailerSuperscript{tt}
\def\TrailerRearAxleEy{e_y^{\TrailerSuperscript}}
\def\TrailerRearAxleEyBar{{\bar e}_y^{\TrailerSuperscript}}
\def\TrailerRearAxleEyHat{{\hat e}_y^{\TrailerSuperscript}}

\def\BusState{z}
\def\ArticulatedState{z}

\def\Beta1{\beta_{1}}

\def\DiscretizationSamplingDistance{\Delta s}
\def\linearizationReferenceVariablesS{ \mathbf{\bar{s}} }
\def\linearizationReferenceVariablesEy{ \mathbf{\bar{e}_y} }
\def\linearizationReferenceVariablesEpsi{ \mathbf{\bar{e}_\psi} }
\def\linearizationReferenceVariablesBetaTwo{ \mathbf{\bar{\beta}_1} }
\def\linearizationReferenceVariablesU{ \mathbf{\bar{u}} }
\def\linearizationReferenceVariablesK{ \mathbf{\bar{\kappa}} }

\section{Introduction}

Driving heavy-duty vehicles is a difficult task that requires expertise and special driver education.
The additional difficulties experienced by drivers when controlling these vehicles translate into further challenges for autonomous driving systems to tackle.
Even though motion planning for autonomous vehicles has been the subject of extensive research efforts, most of its focus has been on passenger cars~\cite{Katrakazas:2015:Survey,Frazzoli:2016:Survey}.
As a result, fundamental problems that affect buses and articulated vehicles, but not passenger cars, have been left unanswered.

The long dimensions of buses are a major challenge for traditional motion planning frameworks.
To be able to plan for such vehicles the work in~\cite{Oliveira:2019:BusDriving} introduces a new environment classification scheme, as well as the formulation of new optimization objectives.
Nevertheless, the planned paths still result in the bus driving unnecessarily close to the road boundaries.

In the tractor-trailer case, the presence of multiple vehicle bodies requires that both bodies are centered simultaneously.
This conflicting objective introduces a trade-off that weighs the importance of centering the tractor, against the importance of centering the trailer.
To tune this parameter, the work in~\cite{Oliveira:2020:OptimizationBased} requires offline time-consuming computations that do not generalize for all vehicle and road combinations.

\begin{figure}[t!]
\centering
\includegraphics[width=1.0\columnwidth]{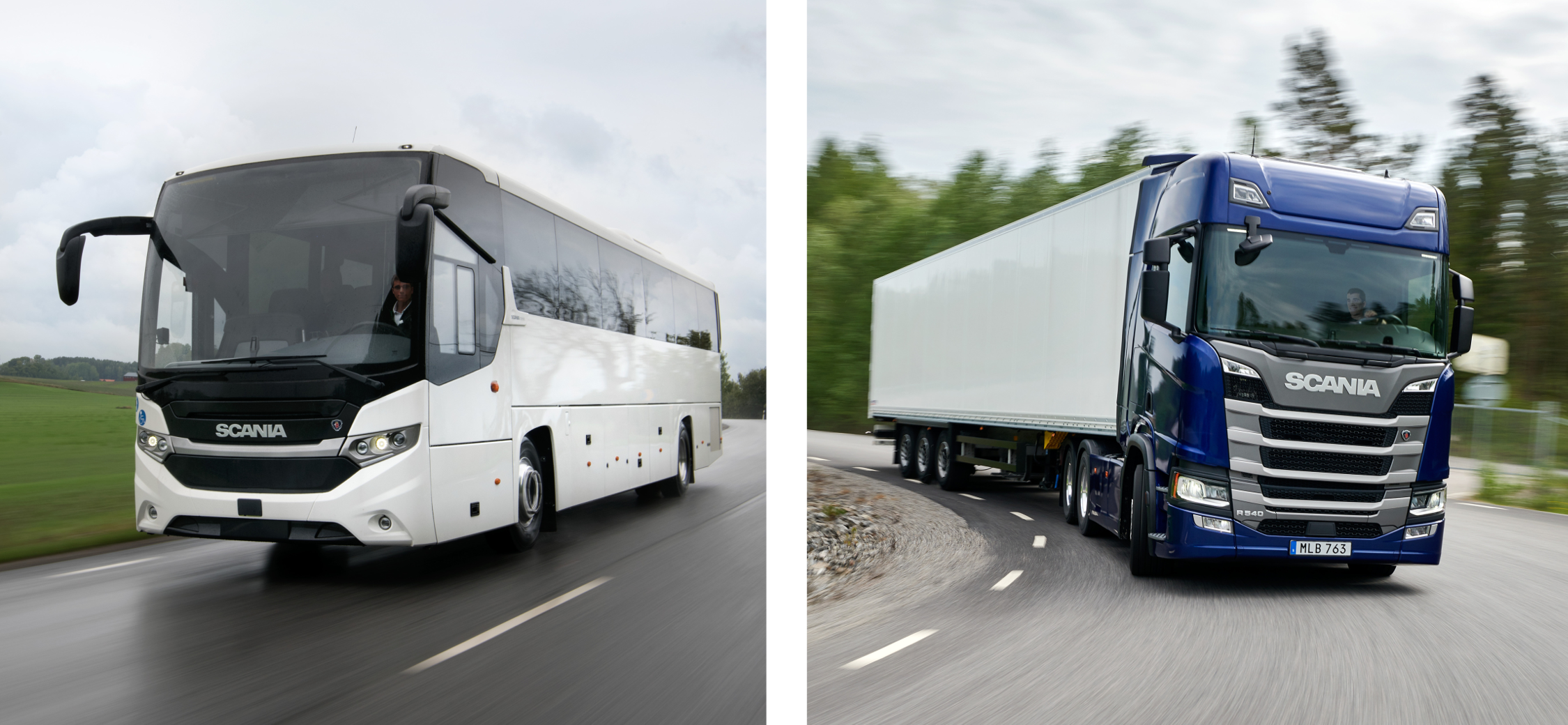}
\caption{\small{A bus (left), with a considerable vehicle length, and a tractor-trailer (right), consisting of two vehicle bodies, are examples of heavy-duty vehicles studied in this work. The long vehicle dimensions, or the presence of multiple vehicle bodies, introduce novel challenges covered in this work. (courtesy of Scania CV)}}
\label{fig:truck-trailer-photo}
\vspace{-15pt}
\end{figure}

In this work, we extend the framework developed in~\cite{Oliveira:2020:OptimizationBased} by deriving, via geometric arguments, the optimal weighting parameter, \textit{i.e.}, the optimal tradeoff between tractor centering and trailer centering.
The derived parameter can then be used in the numerical optimization formulation, resulting in planned solutions that are optimal according to performance metrics measuring the area swept by the vehicle.
Analogously, we derive an optimal weighting parameter for the bus case, this time, corresponding to the optimal tradeoff between rear axle centering and front axle centering.
Using the derived optimal weighting parameter leads to improvements upon the results obtained in~\cite{Oliveira:2019:BusDriving}.

Summarizing, the contributions of this work are:
\begin{itemize}
\item Geometric derivation of optimal driving objectives, with respect to the area swept by the vehicles, suitable for online computation;
\item Development of a unified framework targeting both long vehicles, such as buses, as well as multi-body vehicles, such as tractor-trailers;
\item Comparison with recent works on the same topic~\cite{Oliveira:2020:OptimizationBased,Oliveira:2019:BusDriving}, showing significant performance improvements.
\end{itemize}

The remainder of this paper is structured as follows:~\cref{sec:related_work} sumarizes related work on the motion planning topic.
\Cref{sec:motion_planning} introduces the vehicle models of the bus and the tractor-trailer, and formulates the motion planning problem as a numerical optimization problem.
\Cref{sec:optimal_optimization_objective} presents the geometrical derivation of the driving objectives to be used in the numerical optimization.
\Cref{sec:results} presents relevant simulation results and shows the benefits of the proposed approach when compared to previous methods.
We give final remarks and propose directions for future work in~\cref{sec:conclusions}.

\section{Related Work}

\label{sec:related_work}

The motion planning problem has been the subject of intensive research in the field of robotics.
In order to deal with complex vehicle dynamics~\cite{LaValle:2001:Randomized} proposes Rapidly-exploring Random Trees.
The work in~\cite{Pivtoraiko:2009:StateLattices} proposes instead a special discretization of the search space that is compliant with the vehicle motion capabilities.
Despite good performance when considering unstructured driving environments, these algorithms require specific adaptations to make them suitable for on-road driving~\cite{Kuwata:2008:UrbanDrivingRRT,McNaughton:2011:Conformal}.

In recent years, numerical optimization has emerged as a promising approach to motion planning and control~\cite{Gotte:2016:ARealTimeCapable,Tram:2019:Learning,Svensson:2019:Limits,Ziegler:2014:PlanningForBertha,Lima:2018:Experimental}, due to the broader availability of numerical solvers~\cite{Stellato:2017:Osqp,Ferreau:2014:Qpoases}, as well as increasing computational power available in automotive components.
Numerical optimization approaches benefit from structured driving environments, such as on-road scenarios, and can outperform other existing methods in terms of smoothness and optimality~\cite{Ziegler:2014:PlanningForBertha}.

Research in motion planning for heavy-duty vehicles have mostly considered off-road scenarios~\cite{Ljungqvist:2019:Path,Evestedt:2016:CLRRT,Li:2019:Trajectory}, semi-structured scenarios~\cite{Lamiraux:2005:A380}, or roads with low curvature~\cite{Van:2015:Real}, leaving the challenges of on-road driving opened. 
However, recently proposed works~\cite{Oliveira:2020:OptimizationBased,Oliveira:2019:BusDriving} study the specific problems of heavy-duty vehicles driving on urban roads.

In~\cite{Oliveira:2019:BusDriving}, the authors study the motion planning problem for buses and identify shortcomings in current motion planning frameworks.
A new environment classification scheme together with a new formulation of optimization objectives, increase the maneuverability and safety of buses driving in urban scenarios.
However, the approach only penalizes if the vehicle exits the road boundaries, and therefore planned paths often result in the bus driving on the border of the road boundaries.

The work in~\cite{Oliveira:2020:OptimizationBased} focuses on articulated vehicles and on the problem of how to best drive multi-body vehicles, where centering both vehicle bodies at the same time is a conflicting objective.
The authors propose an optimization objective that is a compromise between the minimization of the area swept by the vehicle, and the feasibility of online computations.
However, the optimization objective includes a tuning parameter used to trade-off between centering the tractor and the trailer on the road.
To achieve the best performance, this parameter has to be properly tuned which can be a time-consuming process.

This paper improves upon the works~\cite{Oliveira:2020:OptimizationBased,Oliveira:2019:BusDriving} by introducing a unified framework targeted for buses and tractor-trailer vehicles.
Our proposed framework improves the driving behavior of buses, by centering their whole body on the lane, thus avoiding the problem of driving too close to road boundaries, as seen in~\cite{Oliveira:2019:BusDriving}.
Moreover, the proposed framework provides a geometric way of computing the tuning parameter presented in~\cite{Oliveira:2020:OptimizationBased}, allowing it to be adapted online to the current road the vehicle is driving in.

\section{Motion Planning Framework}
\label{sec:motion_planning}

This section presents the proposed on-road path planning framework. 
First, the vehicle models for the bus and the tractor-trailer are introduced.
We then formulate the on-road path planning problem as an optimal control problem.

\subsection{Road-aligned bus model}
The vehicles are modeled in a road-aligned frame, which describes the evolution of the vehicles' states in terms of deviation from, and progression along, a geometric reference path $\gamma(\cdot)$. 
The reference path is parametrized in $s$ which corresponds to the distance traveled along the path, and the shape of the path is characterized by a bounded and continuous curvature $\kappa_\gamma(s)$. In this work, the reference path $\gamma(\cdot)$ represents the center of the vehicle's drive lane, but could also be computed by a global path planner.

\begin{figure}
  \centering 
  \resizebox {0.99\columnwidth} {!} {
  \begin{tikzpicture}[scale=1]   
  \def\relativepathtikzfigure{figures/tikz/}
  \input{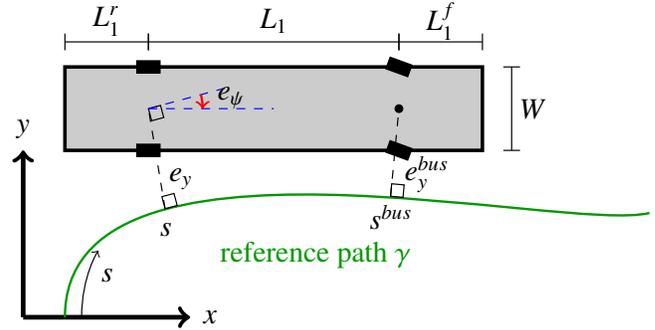}
  \end{tikzpicture}
  }
  \caption{An illustration of the bus in the road-aligned frame and definitions of relevant geometric lengths and
  		vehicle states.}
  \label{fig:bus_road_aligned}
  \vspace{-15pt}
\end{figure}

The bus in the road-aligned coordinate frame is schematically illustrated in Fig.~\ref{fig:bus_road_aligned}. The wheelbase of the bus is denoted by $L_1$ and its width by $W$, whereas $L_1^r$ and $L_1^f$ denote the lengths of the bus rear and front overhangs.      
The vehicle state $s$ represents the distance traveled by the rear axle of the bus along the reference path $\gamma$, whereas $e_y$ is lateral displacement of the bus rear axle with respect to the reference path and $e_\psi$ is the orientation error of the bus with respect to the reference path's tangent. 

The vehicle model is given by~\cite{Gao:2012:Spatial}:
\begin{equation}
\label{eq:temporal_model_bus}
\begin{aligned}
\dot{s} &= v\frac{\cos( e_\psi )}{1 - e_y\kappa_\gamma(s)}, \\
\dot{e}_y &= v\sin( e_\psi ), \\ 
\dot{e}_\psi &= v\left(\kappa - \frac{\kappa_\gamma(s)\cos( e_\psi )}{1-e_y\kappa_\gamma(s)}\right), 
\end{aligned}
\end{equation}
where $\dot{(\cdot)} = \nicefrac{\text{d} (\cdot)}{\text{d}t}$ and the control input $\kappa$ is curvature of the bus.
The relationship between the steering angle of the bus $\phi$ and its curvature is $\kappa = \nicefrac{ \tan( \phi )}{ L_1 }$. 

By restricting the attention to forward motion $v>0$ and employing time scaling with $\dot s>0$, the temporal model of~\cref{eq:temporal_model_bus} can be converted to an equivalent spatial model~\cite{Gao:2012:Spatial}:
\begin{equation}
\label{eq:spatial_model_bus}
\begin{aligned}
e'_y &= \left(1 - e_y\kappa_\gamma\right)\tan( e_\psi ), \\ 
e'_\psi &= \frac{1 - e_y\kappa_\gamma}{\cos( e_\psi )}\kappa - \kappa_\gamma,
\end{aligned}
\end{equation}
where $(\cdot)'=\nicefrac{\text d(\cdot)}{\text ds}$.

\Cref{eq:spatial_model_bus} describes the behavior of the lateral and orientation error of the bus rear axle. 
However, it does not contain information of the lateral error of the bus front axle $\BusFrontAxleEy$ with respect to the reference path $\gamma(\cdot)$. 
To center the whole vehicle body around the reference path, it is desired to represent this auxiliary state $\BusFrontAxleEy$ as a function of \mbox{$[e_y \hspace{5pt} e_\psi]^T$}. 
Except for the case of a straight nominal path, this relationship cannot be written in a purely algebraic form, as it involves a line integral~\cite{altafini:2003OffTracking}. 

In order to be able to consider the lateral error of the bus front axle~$\BusFrontAxleEy$, we introduce its approximation $\BusFrontAxleEyHat$.
Similar to~\cite{Oliveira:2020:OptimizationBased}, it is possible to numerically compute an approximate relationship of $\BusFrontAxleEy$ which depends linearly on the states \mbox{$[e_y \hspace{5pt} e_\psi]^T$}.
Given a linearization point \mbox{$[\bar s\hspace{5pt} \bar e_y\hspace{5pt} \bar e_\psi]^T$}, using finite differences and by iteratively projecting the bus front axle to the reference path, it is possible to compute a linear model for the lateral error of the bus front axle $\BusFrontAxleEy$ as a function of \mbox{$[e_y \hspace{5pt} e_\psi]^T$} that is given by
\begin{align}
\label{eq:partial_derivatives_approx_bus}
\BusFrontAxleEyHat = \BusFrontAxleEyBar + \frac{\partial \BusFrontAxleEy}{\partial e_{y}}(e_y-\bar e_y) + \frac{\partial \BusFrontAxleEy}{\partial e_{\psi}}(e_\psi-\bar e_\psi),
\end{align} 
where the partial derivatives $\frac{\partial \BusFrontAxleEy}{\partial e_{y}}$ and $\frac{\partial \BusFrontAxleEy}{\partial e_{\psi}}$ are computed numerically (see~\cite{Oliveira:2020:OptimizationBased} for details). 

As in~\cite{Oliveira:2019:BusDriving}, the spatial model is discretized to make it suitable for numerical optimization purposes.
Given a path sampling distance $\DiscretizationSamplingDistance$, the reference path is discretized along its length resulting in $\{s_i\}^{N}_{i=0}$ and $\{\kappa_\gamma(s_i)\}^{N}_{i=0}$, where $s_i = i\DiscretizationSamplingDistance$. By defining the state vector for the bus as $\BusState_{\BusSuperscript} = [e_{y}\hspace{5pt} e_{\psi} \hspace{5pt} \BusFrontAxleEy]^T$ and using Euler-forward discretization, a discrete-time nonlinear model of~\cref{eq:spatial_model_bus} and \cref{eq:partial_derivatives_approx_bus} is obtained which can be represented compactly as 
\begin{align}
\label{eq:model:bus}
\BusState_{\BusSuperscript,i+1}=f_{\BusSuperscript}(\BusState_{\BusSuperscript,i},\kappa_i).
\end{align}

\subsection{Road-aligned tractor-trailer model}
\begin{figure}[t!]
	\centering 
	\resizebox {0.99\columnwidth} {!} {
		\begin{tikzpicture}[scale=1]   
		\def\relativepathtikzfigure{figures/tikz/}
		\input{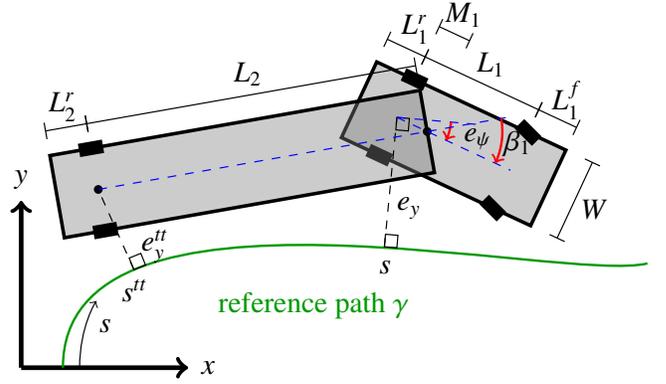}
		\end{tikzpicture}
	}
	\caption{An illustration of the tractor-trailer vehicle in the road-aligned frame and definitions of relevant geometric lengths and vehicle states.}
	\label{fig:tractor_trailer_road_aligned}
	\vspace{-15pt}
\end{figure}

The tractor-trailer vehicle in the road-aligned coordinate frame is illustrated in~\cref{fig:tractor_trailer_road_aligned}. The geometric lengths for the tractor are defined analogously to the bus case and its kinematics modeled accordingly.  
The length $L_2$ is the distance between the trailer's axle and the hitch connection at the tractor, $L_2^r$ is the trailer's rear overhang, and $M_1$ is the signed hitching offset at the tractor. This hitching offset is negative if the hitch connecting is in front of the tractor's rear axle and positive otherwise.
To model the tractor-trailer vehicle's kinematics, one needs to additionally consider state $\Beta1$, the joint angle between the tractor and the trailer.
Its temporal model is given by~\cite{Ljungqvist:2019:Path}:
\begin{equation}
\label{eq:temporal_model_beta_2}
\begin{aligned}
\dot{\Beta1 } &= v \left( \kappa - \frac{\sin( \Beta1 )}{\DTra} + \frac{\DHit}{\DTra} \cos( \Beta1 )\kappa \right),
\end{aligned}
\end{equation}
and as in~\cref{eq:spatial_model_bus}, the equivalent spatial model is
\begin{equation}
\label{eq:spatial_model_beta_2}
\begin{aligned}
\Beta1 ' &= \frac{1 - e_y\kappa_\gamma}{\cos( e_\psi )}\left( \kappa- \frac{\sin( \Beta1 )}{\DTra} +  \frac{\DHit}{\DTra}\cos( \Beta1 )\kappa \right).
\end{aligned}
\end{equation}

Similarly to the bus case,the models in~\cref{eq:spatial_model_bus} and~\cref{eq:spatial_model_beta_2} only provide information about the axle of the tractor, and as such, there is no explicit information regarding the axle of the trailer's lateral error $\TrailerRearAxleEy$ with respect to the reference path $\gamma(\cdot)$.
As no closed-form expression exists to express $\TrailerRearAxleEy$ as a function of \mbox{$[e_y \hspace{5pt} e_\psi \hspace{5pt} \beta_1]^T$} for paths with nonzero curvature, we compute an approximation $\TrailerRearAxleEyHat$ using the techniques presented in~\cite{Oliveira:2020:OptimizationBased}.
Given a working point \mbox{$[\bar s \hspace{5pt} \bar e_y \hspace{5pt} \bar e_\psi \hspace{5pt} \bar\beta_1]^T$}, using finite differences and by iteratively projecting the trailer's axle to the reference path $\gamma(\cdot)$ a linear model of $\TrailerRearAxleEy$ as a function of \mbox{$[e_y \hspace{5pt} e_\psi \hspace{5pt} \beta_1]^T$} is obtained
\begin{align}
\label{eq:partial_derivatives_approx_trailer}
\TrailerRearAxleEyHat = &~\TrailerRearAxleEyBar + \frac{\partial \TrailerRearAxleEy}{\partial e_{y}}(e_y-\bar e_y)  \nonumber \\
&+ \frac{\partial \TrailerRearAxleEy}{\partial e_{\psi}}(e_\psi-\bar e_\psi) + \frac{\partial \TrailerRearAxleEy}{\partial \beta_1}(\beta_1-\bar \beta_1),
\end{align} 
where the partial derivatives $\frac{\partial \TrailerRearAxleEy}{\partial  e_{y}}$, $\frac{\partial \TrailerRearAxleEy}{\partial e_{\psi}}$ and $\frac{\partial \TrailerRearAxleEy}{\partial \beta_1}$ are computed numerically (see~\cite{Oliveira:2020:OptimizationBased} for details). 

We define the state vector as $z_{\TrailerSuperscript} = [e_{y}\hspace{5pt} e_{\psi} \hspace{5pt} \Beta1 \hspace{5pt} \TrailerRearAxleEy]^T$. 
As in the bus case, the reference path is discretized and by performing Euler forward discretization, a discrete-time nonlinear model of the tractor-trailer vehicle~\cref{eq:spatial_model_bus},~\cref{eq:spatial_model_beta_2}, and~\cref{eq:partial_derivatives_approx_trailer} is obtained that is represented compactly as 
\begin{align}
z_{\TrailerSuperscript,i+1}=f_{\TrailerSuperscript}(z_{\TrailerSuperscript,i},\kappa_i).
\label{eq:model:tractor-trailer}
\end{align}

\subsection{Numerical Optimization Formulation}

The on-road path planning problem for the bus (${j=\BusSuperscript}$) and the tractor-trailer vehicle (${j=\TrailerSuperscript}$) are uniformly formulated as the following nonlinear programming (NLP) problem:
\def\OptSpace{\quad}
\begin{subequations}
	\label{eq:optimization_problem}
	\begin{align} 
	\underset{\bm \kappa}{\text{minimize}} & \OptSpace \omega_\kappa J_{\kappa}( \bm \kappa ) + J_{j}( \bm e_y, \bm e_{y}^j ) \label{eq:optimization_objective}\\
	\text{subject to} & \hspace{6pt} z_{j,i+1} = f_j( z_{j,i}, \kappa_i ), \; i \in \{0, ..., N-1\}, \label{eq:constraint_1}\\
	& \OptSpace z_{j,0} = z_{\mathrm{start}}, \; \kappa_0 = \kappa_{\mathrm{start}}, \label{eq:constraint_2}\\
	& \OptSpace p_{e_y}^{\text{obst,s}} \leq g_j( z_{j,i} ), \; i \in \{1, ..., N\}, \label{eq:constraint_3}\\
	& \OptSpace |\kappa_i| \leq \kappa_{\text{max}}, \; i \in \{1, ..., N-1\}, \label{eq:constraint_4}\\
	&\OptSpace |\kappa_i - \kappa_{i-1}| \leq \kappa'_{\text{max}},\; i \in \{1, ..., N-1\}, \label{eq:constraint_5}
	\end{align} 
\end{subequations}
where $\bm e_y = [e_{y,1}~\allowbreak \ldots~\allowbreak e_{y,N}]^{T} \in \mathbb{R}^N$ is the sequence of predicted lateral errors,  $\bm e_{y}^j = [e_{y,0}^j~\allowbreak \ldots~\allowbreak e_{y,N}^j]^{T} \in \mathbb{R}^N$ is the sequence of predicted auxiliary lateral errors and $\bm \kappa = [\kappa_{0}~\allowbreak \kappa_{1}~\allowbreak \ldots~\allowbreak \kappa_{N-1}]^{T} \in \mathbb{R}^N$ is the sequence of vehicle curvatures, corresponding to the commanded control inputs.
The equality constraint in~\cref{eq:constraint_1} corresponds to the vehicle model, where $j=\BusSuperscript$ implies that the bus model~\cref{eq:model:bus} is used and $j=\TrailerSuperscript$ that the tractor-trailer model~\cref{eq:model:tractor-trailer} is used. The constraints in~\cref{eq:constraint_2} are the initial constraints on vehicle's state and curvature. 
The planned paths ensure collision avoidance and keep the vehicle inside of the road limits through constraint~\cref{eq:constraint_3}, where the techniques presented in~\cite{Oliveira:2019:BusDriving} are used. 
The curvature limitations on the tractor and the bus are modeled in~\cref{eq:constraint_4} and~\cref{eq:constraint_5}, where $\kappa_{\text{max}}$ and $\kappa'_{\text{max}}$ are the maximum curvature and curvature rate, respectively.
 
The optimization objective~\cref{eq:optimization_objective} is composed of two
terms.
The term $J_\kappa$ penalizes curvature control inputs and is in this work selected as $J_\kappa(\bm \kappa) = \sum_{i=1}^{N-1}(\kappa_i-\kappa_{i-1})^2$ to promote a smooth curvature profile.
That is, $J_\kappa(\bm \kappa)=0$ if and only if the curvature profile is constant along the entire prediction horizon.
The weight $\omega_\kappa$ determines the importance of driving in a smooth and comfortable manner.

The term $J_j$ penalizes quantities related to the vehicle's lateral errors and is defined as
\begin{align}\label{eq:off_track_objective}
J_{j}( \bm e_y, \bm e_{y}^j ) = \sum_{i=1}^{N}(K_{j,i} e_{y,i} + e_{y,i}^j)^2,
\end{align}  
where $K_{j,i}>0$ is a design parameter. Recall that $e_y$ and $e_{y}^j$ are signed lateral errors, which implies that it is possible that $J_{j}=0$ even though $\bm e_y$ and $\bm e_{y}^j$ are nonzero. This property will be exploited in the next section, where geometric techniques are employed to select $K_{j,i}$ optimally to promote a certain driving behavior.  

To solve the NLP problem~\cref{eq:optimization_problem}, the Sequential Quadratic Programming (SQP) approach presented in~\cite{Oliveira:2019:BusDriving} is used.
At each SQP iteration a Quadratic Programming (QP) problem is constructed, where the vehicle model~\cref{eq:constraint_1} and the collision avoidance constraint~\cref{eq:constraint_3} are linearized around the solution of the previous iteration using a first order Taylor-series expansion.
Moreover, around the previous solution, the linear model for the sequence of auxiliary lateral error $\bm{e_{y}^j}$ is obtained using the approximation~\cref{eq:partial_derivatives_approx_bus} for the bus ($\bm{e_{y}^j}=\bm \BusFrontAxleEy$) and \cref{eq:partial_derivatives_approx_trailer} for the tractor-trailer vehicle ($\bm{e_{y}^j}=\bm \TrailerRearAxleEy$).

\section{Optimal Driving Behavior}
\label{sec:optimal_optimization_objective}

In this section, a desired driving behavior is proposed that accounts for the challenges related to long and multi-body vehicles. 
Based on the desired driving behavior, the optimization objective related to the signed lateral errors~\eqref{eq:off_track_objective} is tuned using geometric conditions. With the proposed optimization objective and design strategy, the result is that the optimal stationary solution to~\eqref{eq:optimization_problem} on roads with constant curvature yields exactly the desired driving behavior.

\subsection{Desired driving behavior}
\label{subsec:desired_driving_behavior}

As presented in~\cite{Oliveira:2020:OptimizationBased}, the formulation of optimization objectives for long and multi-body vehicle is non-trivial.
In fact, due to limited computation time, the optimization objective used in motion planners is often a combination of simple mathematical expressions that favor motion plans making the vehicle behave well according to a desired performance metric. 

When driving vehicles with large dimensions, such as buses or tractor-trailer vehicles, centering one of the vehicle's axles on the center of the road does not suffice to center the whole vehicle on the road. 
Instead, one needs to take particular attention to the whole vehicle body to ensure that all of it is kept as close as possible to the center of the road.

\begin{figure}
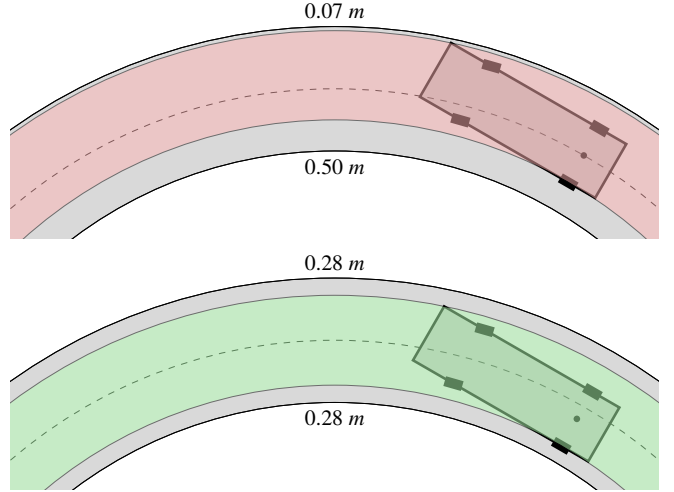

  \centering
	\def\relativepathtikzfigure{figures/tikz/}
	\def\VehicleFrontLength{0.75}
	\def\VehicleAxleLength{2.0}
	\resizebox {0.99\columnwidth} {!} {
  	\begin{tikzpicture}
    \input{figures/tikz/optimal_driving_behavior_bad}
	\end{tikzpicture}
	}
	\resizebox {0.99\columnwidth} {!} {
	\begin{tikzpicture}
    \input{figures/tikz/optimal_driving_behavior_good}
	\end{tikzpicture}
	}
  \caption{The desired behavior is defined as that which achieves the best centering of the vehicle swept area. Top: The vehicle has a rear axle that follows the center of the road, shown as a dotted line. Its swept area, shown in red, tends to the right side of the road. Bottom: The vehicle in has a swept area, shown in green, that is equally distant to both the left and right limits of the road. This desired driving behavior is possible because the rear axle does not follow the center of the road.}
  \label{fig:optimal_behavior_definition}
  \vspace{-15pt}
\end{figure}

As a vehicle progresses along the road, it leaves a trail of its swept area.
This area corresponds to the total covered area that the vehicle's body (or bodies) has occupied while driving along the road.
We define that the whole vehicle body is centered if the maximum extent to which its swept area extends to the left and to right of the center of the road are equal.
The desired driving behavior is shown in~\cref{fig:optimal_behavior_definition}.

In this figure, the red vehicle is driving with its rear axle on the center of the road.
As a result, the area swept by its body, as it progresses along the road, does not have an equal distance to the left and right boundaries of the road. 
On the contrary, the green vehicle has a more desirable driving behavior.
Even though its rear axle is not centered on the road, the spread of its swept area is at equal distances to the left and the right boundaries of the road.

We thus define the desired driving behavior as that resulting in the swept area being equally distant to the left and right boundaries of the road.
With this desired behavior defined, we turn to the problem of designing $K_{j,i}$ in~\eqref{eq:off_track_objective} such that this behavior is obtained for the bus and tractor-trailer vehicles.

\subsection{Derivation for the bus case}
\label{subsec:derivation_bus}

\begin{figure}
	\centering 
	\resizebox {0.99\columnwidth} {!} {
		\begin{tikzpicture}[scale=1]   
		\def\relativepathtikzfigure{figures/tikz/}
		\input{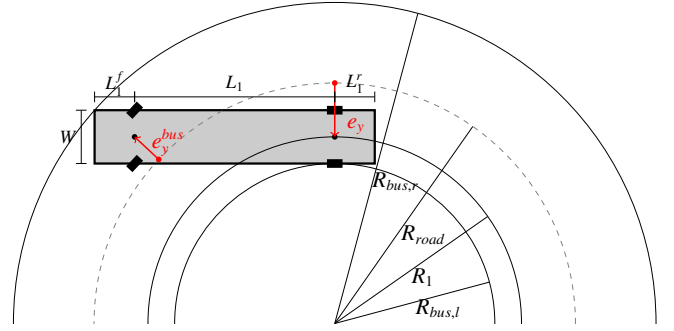}
		\end{tikzpicture}
	}
	\caption{Geometric illustration of optimal road centering for a bus on a counter-clockwise turn with constant radius $R_{road}$.}
	\label{fig:geometric_derivation_bus}
	\vspace{-15pt}
\end{figure}

Fig.~\ref{fig:geometric_derivation_bus} illustrates the scenario of a bus driving along a road with a constant radius $R_{road}$ and achieving the desired driving behavior.
The bus is assumed to drive at a constant curvature $\kappa>0$ which renders in a constant, but unknown, turning radius $R_{1}=1/\kappa$. Since $|\kappa|\leq \kappa_{max}$, the bus turning radius satisfies $|R_{1}|\geq 1/\kappa_{max}$. Without loss of generality, it is further assumed that the bus is driving in a left turn, giving that $R_{1}>0$. The derivations for a right turn is done similarly. 
In a left turn, the swept area is delimited by the radius $R_{\BusSuperscript,l}$ corresponding to the path traveled by the bus rear left wheel, and by the radius $R_{\BusSuperscript,r}$ corresponding to the path traveled by the front right corner of the bus body.
To achieve that the swept area by the vehicle's body is equally spread to the right and to the left of the road center, the following relationship must hold:
\begin{equation}
\label{eq:geometric_bus_1}
R_{road} = \frac{R_{\BusSuperscript,l} + R_{\BusSuperscript,r}}{2}.
\end{equation}
As the turning radius of the bus $R_{1}$ is constant, basic trigonometry gives that the inner and outer radii $R_{\BusSuperscript,l}$ and $R_{\BusSuperscript,r}$ can be represent as:
\begin{align} 
\label{eq:geometric_bus_2}
\begin{split}
R_{\BusSuperscript,r}^2 &= \left(R_{1}+\frac{W}{2}\right)^2 + \left(L_1 + L_1^f\right)^2,\\
R_{\BusSuperscript,l} &= R_{1}-\frac{W}{2},
\end{split}
\end{align} 
where it is assumed that $R_{1}>W/2$. Note that this assumption does not pose any practical restrictions as the minimum turning radius of a bus is typically much larger than half the vehicle's width.
Since $R_{\BusSuperscript,r}>0$ in a left turn, inserting~\eqref{eq:geometric_bus_2} in~\eqref{eq:geometric_bus_1} yields:
\begin{equation}
\label{eq:geometric_bus_3}
R_{road} = \frac{\sqrt{\left(R_{1}+\frac{W}{2}\right)^2 + \left(L_1 + L_1^f\right)^2} + R_{1}-\frac{W}{2}}{2},
\end{equation}
which is a nonlinear equation in the unknown variable $R_{1}$. For roads with radius $R_{road}$ such that $R_1>W/2$, the unique and positive solution to~\eqref{eq:geometric_bus_3} is 
\begin{equation}
\label{eq:geometric_bus_4}
R_{1} = \frac{-\left(L_1 + L_1^f\right)^2 + 4R_{road}^2 + 2WR_{road}}{4R_{road} + 2W}.
\end{equation}
\Cref{eq:geometric_bus_4} gives the optimal turning radius of the bus $R_{1}$ as a function of the road curvature $R_{road}$, which is optimal in the sense that the bus left swept width ($R_{road}-R_{\BusSuperscript,l}$) and right swept width ($R_{\BusSuperscript,r}-R_{road}$) are equal.
This is deemed as the desired behavior as it perfectly centers the area swept by the vehicle around the road center.

From the derived turning radius of the bus $R_{1}$ it is now possible to obtain the constant sign lateral errors $\BusFrontAxleEy$ and $e_{y}$, corresponding to the bus front and rear axle distances to the road center are given by:
\begin{align} 
\label{eq:geometric_bus_5}
\begin{split}
\BusFrontAxleEy &= R_{road} - \sqrt{ L_1^2 + R_{1}^2 },\\
e_{y} &= R_{road} - R_{1},
\end{split}
\end{align}
where $\BusFrontAxleEy<0$ and $e_{y}>0$.  
To make the optimization objective $J_{\BusSuperscript}=0$ at this stationary configuration, we get from~\eqref{eq:off_track_objective} that $K_{\BusSuperscript} e_y + \BusFrontAxleEy=0$ must hold. This condition together with~\eqref{eq:geometric_bus_5} gives the optimal tuning strategy
\begin{align}
\label{eq:tuning_bus}
K_{\BusSuperscript}(R_{road}) = \frac{\sqrt{\rule{0pt}{2ex} L_1^2 + R_{1}^2} - R_{road}}{R_{road}-R_{1}}.
\end{align} 
For the case of a clockwise turn with equal radius, the same geometrically derived tuning of $K_{\BusSuperscript}$ can be used.
With the proposed tuning of $K_{\BusSuperscript}$, and under the assumption of no obstacles or other additional vehicle constraints, the optimization objectives $J_{\BusSuperscript}$ and $J_\kappa$ will obtain their minimum value of zero when the vehicle moves along the road with a constant curvature \mbox{$\kappa=1/R_{1}$}, where $R_{1}$ is given by~\eqref{eq:geometric_bus_4}. 
Using this tuning strategy, the optimization-based path planner is guided towards finding a solution with the desired behavior of having a balanced swept width to the left and to the right of the road center.

\subsection{Derivation for the tractor-trailer case}
\label{subsec:derivation_tractor_trailer}

\begin{figure}[t!]
	\centering
	\resizebox {0.99\columnwidth} {!} {
		\begin{tikzpicture}[scale=1]   
		\def\relativepathtikzfigure{figures/tikz/}
		\input{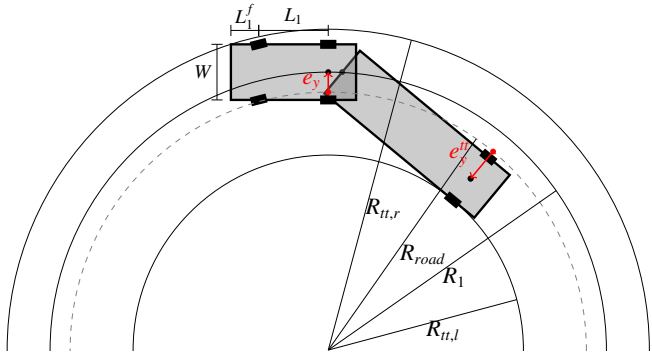}
		\end{tikzpicture}
	}
	\caption{Geometric illustration of the stationary and optimal road centering of a tractor-trailer vehicle around a counter-clockwise turn with constant radius $R_{road}$.}
	\label{fig:tractor_trailer_road_aligned_circular}
	\vspace{-10pt}
\end{figure}

\Cref{fig:tractor_trailer_road_aligned_circular} illustrates the tractor-trailer vehicle driving along a road with a constant radius $R_{road}$ and achieving the desired driving behavior. 
The tractor-trailer vehicle is posed in a stationary circular equilibrium configuration~\eqref{eq:circular_eq_configuration} where a constant curvature of the tractor $\kappa$ corresponds to $\beta_1' = 0$ and
\begin{align}
\label{eq:circular_eq_configuration}
\beta_1 = \left(\arctan\left(\frac{M_1}{R_1}\right) + \arctan\left(\frac{L_2}{R_2}\right)\right),
\end{align} 
where the signed radii $R_1 = 1/\kappa$ and $R_2^2 = R^2_1+M_1^2-L_2^2$.
The swept area by the vehicle's bodies is characterized by radius $R_{\TrailerSuperscript,l}$ corresponding to the path traveled by the rear left wheel of the trailer, and by the radius $R_{\TrailerSuperscript,r}$ corresponding to the path traveled by the front right corner of the tractor's body. 
In analogous to the bus case, to achieve the desired driving behavior, the following relationship must hold:
\begin{align}
\label{eq:geometric_tt_1}
R_{road} = \frac{R_L+ R_R}{2}.
\end{align}   
As the turning radius of of the tractor $R_1=1/\kappa$ and the joint angle $\beta_1$ are both constant, basic trigonometry gives that the radii $R_{\TrailerSuperscript,l}$ and $R_{\TrailerSuperscript,r}$ are given by:
\begin{align} 
\label{eq:geometric_tt_2}
\begin{split}
R_{\TrailerSuperscript,r}^2 &= \left(R_1+W/2\right)^2 + \left(L_1 + L_{1}^f\right)^2,\\
R_{\TrailerSuperscript,l} &= R_{2}-W/2,
\end{split}
\end{align}
where it is assumed that the turning radius of the trailer's axle $R_{2}>W/2$, which is typically true for standard roads. 
Since $R_{\TrailerSuperscript,r}>0$, inserting~\eqref{eq:geometric_tt_2} in~\eqref{eq:geometric_tt_1} gives 
\begin{align}
\label{eq:geometric_tt_3}
2R_{road} =&\sqrt{\rule{0pt}{2ex}R^2_1+M_1^2-L_2^2}-W/2 \nonumber \\ &+ \sqrt{\left(R_1+W/2\right)^2 + \left(L_1 + L_{1}^f\right)^2},
\end{align}
which is a nonlinear equation in the variable $R_1$. The positive solution to~\eqref{eq:geometric_tt_3} can compactly be represented as
\begin{align}
\label{eq:geometric_tt_R1}
R_1 = g(R_{road},W,L_1,L_2,M_1,L_1^f).
\end{align}
Function $g$ is found using MATLAB's symbolic toolbox, however, due to its extensive length, it is not presented in the paper.
It is now possible to compute $R_2$ and also the joint angle $\beta_1$ using~\eqref{eq:circular_eq_configuration}. 
From the derived radii $R_{1}$ and $R_2$, the constant signed lateral errors $\TrailerRearAxleEy$ and $e_{y}$, corresponding to the trailer's axle and the tractor's rear axle distances to the road center are given by:
\begin{align}
\label{eq:geometric_tt_4}
\begin{split}
\bm e_{y} &= R_{\text{road}} - R_1, \\
\bm \TrailerRearAxleEy &= R_{\text{road}} - \sqrt{\rule{0pt}{2ex} R_1^2+M_1^2-L_2^2},
\end{split}
\end{align}
where $e_y<0$ and $\TrailerRearAxleEy>0$. 
In analogous to the bus case, to make the optimization objective $J_{\TrailerSuperscript}=0$ at this stationary configuration, we get from~\eqref{eq:off_track_objective} that $K_{\TrailerSuperscript} e_y + \TrailerRearAxleEy=0$ most hold. This together with~\eqref{eq:geometric_tt_4} gives the optimal tuning
\begin{align}\label{eq:tuning_trailer}
K_{\TrailerSuperscript}(R_{road}) = \frac{R_1 - R_{road}}{R_{road}-\sqrt{\rule{0pt}{2ex} R_1^2+M_1^2-L_2^2}},
\end{align}
where $R_1$ is given in~\eqref{eq:geometric_tt_R1}.
When considering a clockwise turn with equal radius, the same geometrically derived tuning of $K_{\TrailerSuperscript}$ can be used.
With the proposed tuning of $K_{\TrailerSuperscript}$, and under the assumption of no obstacles or other additional vehicle constraints, the optimization objectives $J_{\TrailerSuperscript}$ and $J_\kappa$ will be zero when the tractor-trailer vehicle moves along the road with a constant joint angle~\eqref{eq:circular_eq_configuration} and a constant curvature of the tractor \mbox{$\kappa=1/R_{1}$}, where $R_{1}$ is given by~\eqref{eq:geometric_tt_R1}.
Thus, using this tuning strategy the optimization-based path planner is guided towards finding a solution that achieves the desired behavior of having a balanced swept area of the tractor-trailer bodies to the left and the right of the road center.

\subsection{Roads with varying curvature}

The geometrically derived tuning of $K_{\BusSuperscript}$ and $K_{\TrailerSuperscript}$ can now be used in the path planner's optimization objective~\eqref{eq:off_track_objective}.
If the road has a constant curvature, one can simply define $K_{j,i}$ in~\eqref{eq:off_track_objective} to be equal to the derived $K_j$ in~\cref{subsec:derivation_bus} and~\cref{subsec:derivation_tractor_trailer}.
For the generic driving situation in which the road has a varying curvature, one needs to update $K_{j,i}$ along the planning horizon.
In this work, this is done by updating $K_{j,i}$ at each point along the sampled reference path $\{\gamma(s_i)\}^{N}_{i=0}$ based on its curvature $\kappa_\gamma(s_i)$.
We note that the geometric derivations in~\cref{subsec:derivation_bus} and \cref{subsec:derivation_tractor_trailer} assume a road with constant curvature.
However, as is shown in the next section, using a varying $K_{j,i}$ based on the road curvature results in a behavior that is close to the one expected based on constant curvature assumptions.

\section{Results}
\label{sec:results}

This section presents results showing the advantages of the proposed framework.
The results consider both the bus and tractor-trailer cases and compare them with previous works.
Furthermore, we test the performance of the motion planner in data gathered from real roads.

The results presented are obtained using a laptop computer with an Intel Core i7‐6820 HQ@2.7GHz CPU, with the code implemented in MATLAB.
We use OSQP~\cite{Stellato:2017:Osqp} to solve the SQP iterations of the motion planning problem.
For the vehicle dimensions we use the bus described in~\cite{Oliveira:2019:BusDriving} and the tractor-trailer described in~\cite{Oliveira:2020:OptimizationBased}.

\subsection{Bus in a U-turn}

\def\PlannedPathSizes{0.95}

\def\tikzLegendSize{\large}
\begin{figure}
  \centering 
  \resizebox {0.99\columnwidth} {!} {
    \input{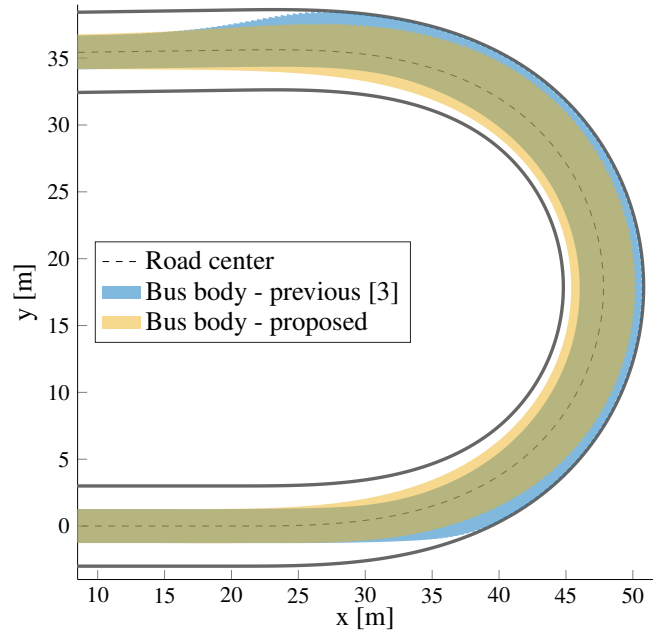}
	}
        \caption{In previous work~\cite{Oliveira:2019:BusDriving}, the bus drives close to the border of the road (blue). However, using the proposed method, the bus is able to center itself on the road (yellow).}
        \label{fig:bus_improvement_planned_path}
        \vspace{-15pt}
\end{figure}

It is noticeable that in previous work~\cite{Oliveira:2019:BusDriving}, driving the bus on a turn results in the vehicle being excessively close to the outer road limits.
This is due to that the optimization objective only considers the lateral position of the rear axle.
Using an optimization objective that considers both the rear and the front axle, and using the derived optimal $K_{j,i}$ results in that the bus drives centered on the road.
\cref{fig:bus_improvement_planned_path} replicates the results presented in~\cite{Oliveira:2019:BusDriving}, where the bus drives on the border of the road, and compares them to the results obtained by our proposed method, where the bus drives centered on the road.

\subsection{Tractor-trailer in a roundabout}

\begin{figure}
  \centering 
    \resizebox {0.99\columnwidth} {!} {
  	\input{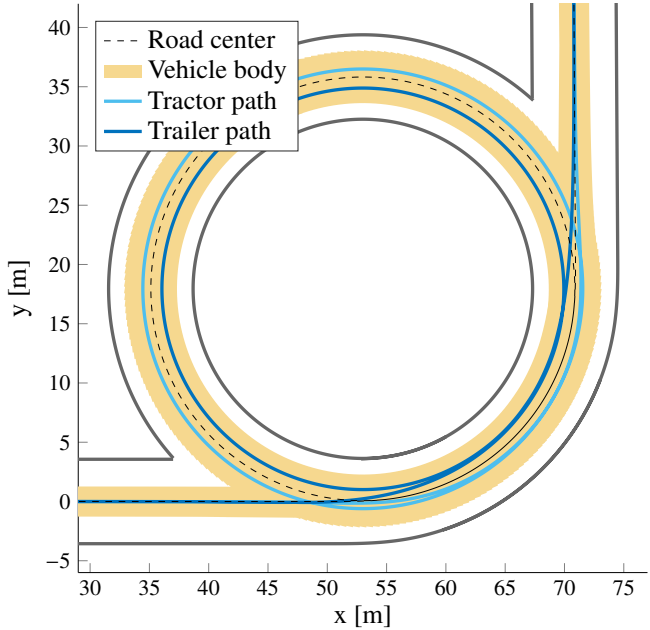}
	}
        \caption{The tractor-trailer vehicle is able to center its whole body as it drives along the roundabout.}
        \label{fig:tractor_trailer_improvement}
\end{figure}

We now consider the tractor-trailer driving in a roundabout scenario as presented in~\cite{Oliveira:2020:OptimizationBased}.
Using the proposed geometric method to obtain the optimal weighting parameter, the planned solution is able to center the vehicle precisely, as shown in~\cref{fig:tractor_trailer_improvement}.
The maximum envelope widths to the left and right of the road center differ only by $0.04$ m as presented in~\cref{fig:tractor_trailer_improvement_envelopes}, and are both very close to the optimally derived envelope width.
A transient behavior can be observed at the entrance and exit of the roundabout, however, for a considerable length of the maneuver, the vehicle is driving according to the geometrically derived optimal stationary behavior.

\def\tikzLegendSize{\large}

\def\tikzLegendSize{\Large}
\begin{figure}
  \centering 
  \resizebox {0.99\columnwidth} {!} {
  	\input{figures/script_03_envelopes.tikz}
	}
        \caption{The proposed motion planner achieves a balanced tractor and trailer centering, where the maximum left and right widths correspond to $2.30$ m and $2.26$ m respectively. These widths are both fairly close to the expected width of $2.27$ m derived geometrically. \message{LaTeX Warning: The width should be quite close to the expected if not, then it probably is the result of a discretization issue! Try to use finer discretization.}}
        \label{fig:tractor_trailer_improvement_envelopes}
\end{figure}

\begin{figure}
  \centering
    \resizebox {0.99\columnwidth} {!} {
  	\input{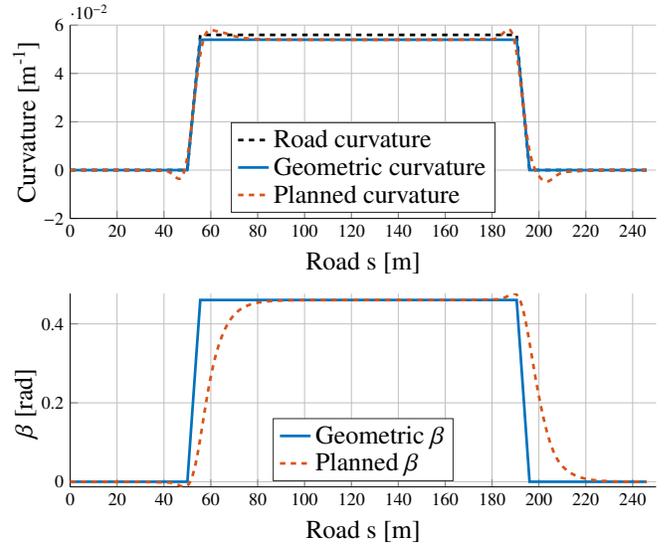}
	}
        \caption{Top: The solution path curvature of the proposed motion planner closely follows the optimal curvature derived using stationarity principles. Bottom: The same is true with respect to the articulation angle $\beta$.}
        \label{fig:tractor_trailer_improvement_curvatures}
        \vspace{-15pt}
\end{figure}

To further validate our geometrical approach,~\cref{fig:tractor_trailer_improvement_curvatures} compares the derived optimal vehicle curvature, obtained at each road distance $s$, with the planned vehicle curvature found by the motion planner.
It can be seen that the planned curvature follows the optimal curvature quite closely, with the exception of transients at the entrance and exit of the roundabout.
\cref{fig:tractor_trailer_improvement_curvatures} also compares the planned and optimal articulation angles $\beta$.
In contrast to the curvature, the articulation angle has a slower response time, as can be seen by the significantly longer transient behavior.

The optimal $K_{\TrailerSuperscript,j}$ values used in this experiment are computed online using~\eqref{eq:tuning_trailer}.
In previous work~\cite{Oliveira:2020:OptimizationBased}, $K_{\TrailerSuperscript,j}$ would have to be computed offline, either by manual tuning, or automatically found by trying out different values and choosing the best.
Both options are quite time consuming, and the results cannot be generalized for different vehicle configurations or road scenarios.
With the proposed method, we are able to compute the optimal $K_{\TrailerSuperscript,j}$ values online, allowing them to dynamically adapt to the current road characteristics.
This represents a significant improvement over the work~\cite{Oliveira:2020:OptimizationBased}.

\subsection{Computational times on real road data}

We run tests on road data obtained from Scania’s test
facilities in Södertälje, Sweden, and measure the computational times of the proposed methods for the bus and tractor-trailer cases.
For both vehicles, we assuming a planning horizon of $100$ m, and a discretization of the reference path of $0.5$ m.
The motion planner is implemented in a receding horizon fashion, where every planned path is executed for the first $5$ m and then a new plan is computed over a shifted horizon.

Problem~\ref{eq:optimization_problem} is solved using an SQP approach~\cite{Oliveira:2019:BusDriving}.
The SQP algorithm can run until convergence of the solution, \textit{i.e.}, until the solution of a given QP is arbitrarily close to the linearization point.
Alternatively, it can run in an Real-Time Iteration (RTI) fashion~\cite{Svensson:2019:Limits}, where only one QP iteration is performed at each planning step.
This is particularly suited for motion planners that operate in a receding horizon, since the linearization of the QP step, will correspond to the previous motion planner solution, therefore resembling an SQP that runs until convergence of the solution.

The computational results for both methods are shown in~\cref{tab:computational_times}.
It has been observed in our experiments that both the SQP and RTI schemes results in almost identical vehicle performances.
With respect to computational times, the RTI scheme is on average twice as fast as the SQP, however its biggest advantage comes from the worst-case planning time, where it is an order of magnitude faster.

\def\TableFontSize{\footnotesize}
\begin{table}
\centering
\caption{\small{Computational times of the proposed method.}}
\begin{tabular}{|c|c|c|c|c|}\hline
\multirow{2}{*}{\TableFontSize Vehicle} & \multicolumn{2}{|c|}{\TableFontSize SQP time} & \multicolumn{2}{|c|}{\TableFontSize RTI time}
\\\cline{2-5}
 & \multicolumn{1}{|c|}{\TableFontSize mean} & \multicolumn{1}{|c|}{\TableFontSize max} & \multicolumn{1}{|c|}{\TableFontSize mean} & \multicolumn{1}{|c|}{\TableFontSize max} \\
\hline
{\TableFontSize Bus} & \TableFontSize $0.079$ s & \TableFontSize $0.742$ s & $0.030$ s & $0.048$ s\\
\hline
{\TableFontSize Tractor-trailer} & \TableFontSize $0.137$ s & \TableFontSize $1.248$ s & \TableFontSize $0.050$ s & \TableFontSize $0.137$ s\\
\hline
\end{tabular} 
\label{tab:computational_times}
\vspace{-10pt}
\end{table}

\section{Conclusions}
\label{sec:conclusions}

We have introduced a framework for designing optimization objectives of motion planners.
The developed approach targets both buses and tractor-trailers, resulting in a unified framework for a large number of possible heavy-duty vehicle configurations.
To design the optimization objective, we define the desired driving behavior to be that resulting in the whole vehicle body driving as centered on the road as possible.
We then use a computationally efficient optimization objective to achieve this complex driving behavior.
The optimization objective formulation is obtained via geometric arguments, being suitable for online computation, allowing for a continuous adaptation of the optimization objective to the current road characteristics.
Our results show significant improvements upon previous works targeting buses and tractor-trailers.
Tests using real road data highlight the capability of the method to tackle real-world scenarios and indicate its computational tractability.

As future work, we will generalize the developed framework to more complex vehicles, such as vehicles with multiple actuated steering axles, and articulated vehicles composed of a tractor, a dolly, and a trailer.
The framework can be readily extended to consider alternative desired driving behaviors besides that of centering the vehicle on the road.
Based on the current traffic situation, it might be beneficial  to plan paths that maximize the distance between the vehicle swept area and oncoming traffic.
To further validate the approach, we plan to implement the proposed methods on real world tests using autonomous heavy-duty vehicles.

\bibliographystyle{ieeetran}
{\footnotesize
\bibliography{references}
}

\end{document}